\documentclass{llncs}
\usepackage{llncsdoc}
\usepackage{amsmath}
\usepackage{amsfonts}
\usepackage{ascmac}
\usepackage{graphicx}
\graphicspath{{figures/}}
\usepackage{bm}
\begin{document}
\newcounter{save}\setcounter{save}{\value{section}}
{\def\addtocontents#1#2{}%
\def\addcontentsline#1#2#3{}%
\def\markboth#1#2{}%
%
\title{Modularity Optimization as a Training Criterion for Graph Neural Networks}

\author{Tsuyoshi Murata \and Naveed Afzal}

\institute{Department of Computer Science, School of Computing Tokyo
Institute of Technology\\ W8-59 2-12-1 Ookayama, Meguro, Tokyo, 152-8552 Japan\\
\email{murata@c.titech.ac.jp\\http://www.net.c.titech.ac.jp/}}

\maketitle
\begin{abstract}
Graph convolution is a recent scalable method for performing deep feature learning on attributed graphs by aggregating local node information over multiple layers. Such layers only consider attribute information of node neighbors in the forward model and do not incorporate knowledge of global network structure in the learning task. In particular, the modularity function provides a convenient source of information about the community structure of networks. In this work we investigate the effect on the quality of learned representations by the incorporation of community structure preservation objectives of networks in the graph convolutional model. We incorporate the objectives in two ways, through an explicit regularization term in the cost function in the output layer and as an additional loss term computed via an auxiliary layer. We report the effect of community structure preserving terms in the graph convolutional architectures. Experimental evaluation on two attributed bibilographic networks showed that the incorporation of the community-preserving objective improves semi-supervised node classification accuracy in the sparse label regime. 
\end{abstract}
\section{Introduction}


\par In recent years, Convolutional Neural Networks (CNNs) \cite{lecun1998gradient} have successfully exploited the statistical regularities on several domains and have shown state-of-the-art results in image classification \cite{krizhevsky2012imagenet}, speech recognition \cite{hinton2012deep} and related tasks. CNNs achieve this by making two important assumptions about the statistical properties of the data, locality and translation invariance. These assumptions are exploited by learning local feature extractors that are shared across the domain. Defining such feature extractors requires clear notions of translation and locality. These are clear on domains where the sample dependencies have a regular Euclidean structure (ie. images, videos, text and speech). However, in many data domains the relationships between the samples is arbitrary and cannot be expressed as a regular Euclidean structure. The fields of social networks analysis, bioinformatics, and information science all involve data with various complex relationships between the entities. These relationships can be represented mathematically with a graph $G(V, E)$ where $V$ is a set of vertices and $E$ is a set of pair-wise relationships between them. Each vertex in $V$ represents a single training sample in $\bf{X}$ and an edge between two vertices can represent some domain-specific relationship. 

\par Starting with \cite{bruna2013spectral}, recent efforts have attempted to generalize the main assumptions of CNNs to arbitrary graph domains. With \cite{defferrard2016convolutional}, we now have scalable formulations of convolutional layers applicable to general graphs, and are denoted in the literature as graph convolutional layers. Given a definition of proximity of graph vertices as defined by simple k-hop distances, the graph convolutional layer extracts local features for each node. These feature extractors are replicated over every node of the graph, effectively limiting the number of learnable parameter to be independent of graph size. This model has been shown to be very effective in various graph-based learning tasks as demonstrated in \cite{kipf-gcn} and \cite{kipf2016variational}.

\par The assumption of locality on networks can be used to learn very useful models but real networks also tend to exhibit various global properties. The field of Network Science \cite{newman2003structure} is generally concerned with studying such high-level properties of networks. An important feature of real networks is the existence of dense clusters or communities where the vertices of a network have a higher density of edges among them that between clusters. \cite{newman2006modularity} defined graph clustering as an optimization problem with an objective function called modularity. Modularity is a global measure of the networks' structure and can thus provide a higher-level view of the network's properties than considering local neighborhood information alone.


\par This work is based on the assumption that injecting additional information about higher-level network structure into a neural architecture trained on an appropriate graph-based learning task can improve performance on that task. Recent attempts \cite{tu2016community} \cite{wang2017community} \cite{yang2016modularity} have successfully integrated community structure information into various graph representation learning methods. Similar work has not yet been attempted for graph convolutional models. In situations where the number of labeled examples for training is very sparse, the model should be able to benefit from leveraging more higher level network structure information as compared to learning from labels alone. In this work, we incorporate community structure information by integrating modularity score optimization into the framework of graph convolutional neural networks. For evaluation of our approach, we explore the case of semi-supervised node classification on graph datasets in the sparse label regime.


\section{Background}

\subsubsection{Community or Meseoscopic Structure}
A graph $G$ can be quantified as an $n \times n$ adjacency matrix $A$ where $n$ is the number of nodes. If node $i$ and $j$ have an edge between them then entry $A_{ij}$ of the matrix is $1$ else it is $0$. 

\par Networks tend to be organized in a higher structural level into clusters. A cluster or community in a network is a group of vertices that has a higher density of edges among nodes within the group but sparser connections to nodes outside. 



\par \cite{newman2006modularity} proposed modularity as a score to measure the goodness of partition of a given network. Statistically, it is the fraction of edges within a group minus the expected fraction for a random graph with same degree distribution, summed over every group. It is the most commonly used objective function for community detection. Modularity optimization provides additional information not normally available from optimizing neighborhood structure alone.
%
%
Given a group assignment Matrix $H$, the score of the partition can be computed as:
\begin{equation}
 Q = tr(\bm{H}^T\bm{B}\bm{H}),
\end{equation}

\noindent
where $ B_{ij} = A_{ij} - \frac{k_i k_j}{2e} $ is the modularity matrix.

\subsection{Graph Neural Network}

The standard convolutional filters are not directly applicable to arbitrary graph structures due to lack of clear notion of translation and ordering of the node neighborhoods.

\subsubsection{Locality and Translational Invariance}
\par A CNN layer is simply a filter of learnable parameters that is convolved on the input pixels of an image. Naturally, these layers possess the properties of locality and translational invariance. The locality property is based on the assumption that the statistical dependence of pixels in an image is inversely dependent on the distance between them. In practice, this is exploited by making the filtering step consider only pixel values in a fixed local neighborhood which is defined by the size of the filter. The second assumption is that of spatial model invariance, which is based on the observation that the identity of an object in an image does not change regardless of its translation in the image. The CNN filter exploits this property by sharing the same set of weights for every image patch. Through these assumptions the number of parameters of the convolutional layer can be made independent of the input size. A suitable definition of convolutional layers on graphs should possess these two properties.

\begin{figure}[h]
	\centering
	\includegraphics[width=0.6\textwidth]{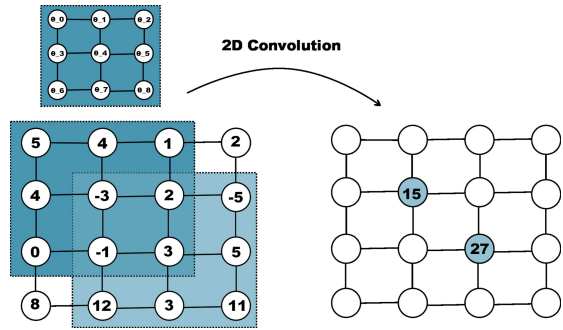}
	\caption{Standard Convolutional Layer}
\end{figure}



\subsubsection{Polynomial Graph Convolutional Filters}
\par \cite{defferrard2016convolutional} defined localized graph Convolutional filters directly computed in the spatial domain. The filter is defined as a polynomial function of the graph Laplacian matrix. Besides a simple polynomial formulation, a model computed via Chebyshev polynomials \cite{hammond2011wavelets} has also been defined in the same work. These formulations have computational complexity of order equal to number of graph edges due to sparse multiplications. An important property of polynomial convolutions is that the locality of the filter is equal to the order of the polynomial. \cite{kipf-gcn} defined a simplified version of this by restricting the model to first-order filters. In this document we will refer to Kipf's model as GCN and Defferard's Chebyshev formulation as ChebNet. These class of models posses both locality and translation invariance properties of graph convolutions and have scalable computation time. The model described in this paper are also based on the polynomial graph convolutions of Defferard and Kipf and will be described in greater detail.

\subsubsection{Regularizing Deep Models via Graph Structure}
\par \cite{weston2012deep} described an approach for incorporating graph structural information directly into any arbitrary deep architecture during training. The structural information is incorporated as an unsupervised loss term directly computed over the representations in any layer of the deep architecture. The final loss function is then composed of two parts, the supervised loss over labeled examples and the unsupervised loss over all example regardless of labels. The domain structure can be incorporated in three ways as illustrated in Figure \ref{fig:weston}, either by (a) directly adding the unsupervised term to the output layer loss of the architecture, (b) by computing the loss directly on the representations on any non-output layer, or by (c) attaching a new auxiliary feed-forward layer to the architecture and computing the loss on the output of that layer. In this work, the modularity optimization term is incorporated into a graph convolutional architecture based on these techniques. The resulting models will be described in detail in the next section.

\begin{figure}[h]
	\centering
	\includegraphics[width=0.9\textwidth]{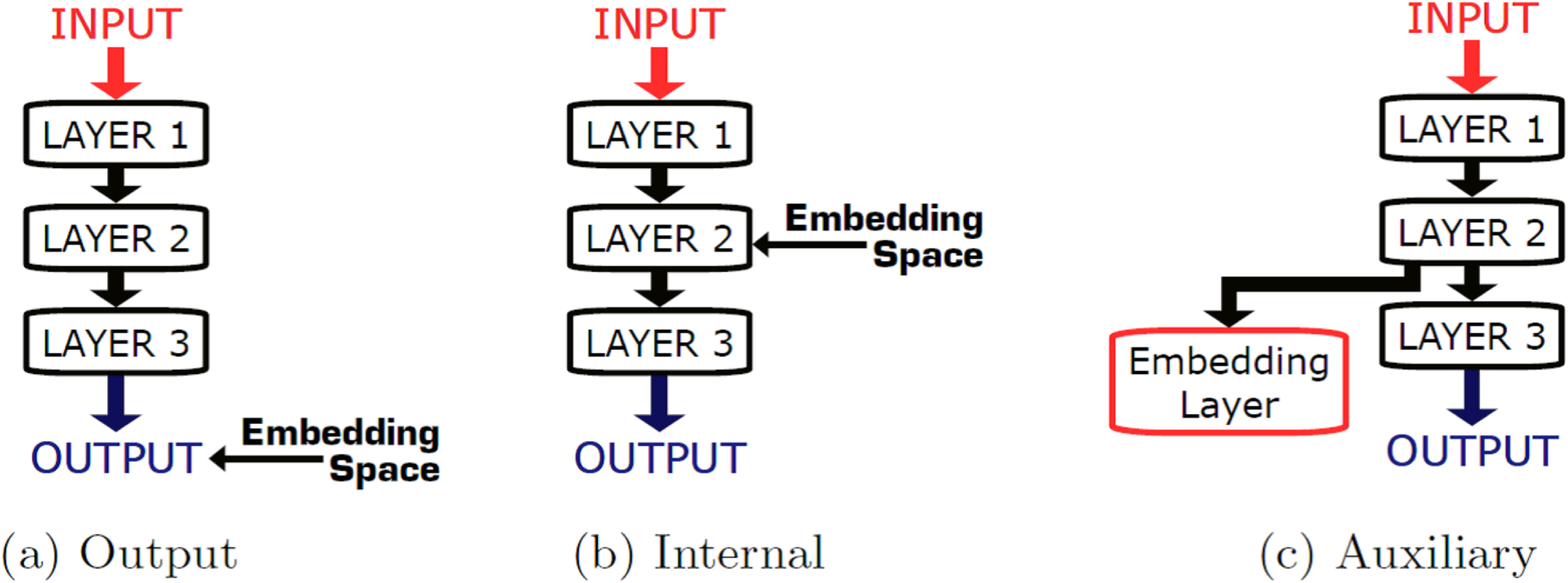}
	\caption{Semi-Supervised Embedding \cite{weston2012deep}}
	\label{fig:weston}
\end{figure}

\section{Model Description}


\subsection{Encoding Model}
This section will describe the encoding layers used in the experiments. The inputs are the attribute and adjacency matrices of the graph to be encoded and the output is a latent representation $Z$ of any arbitrary dimensionality. The latent representations depend on the weights and the biases of the encoder model which can be trained using gradient descent. The encoders can also be made arbitrarily deeper by stacking additional layers.
Given a Graph Convolutional layer with $C$ input channels (features) and $F$ output features:

GCN: $W^t$ should be a $C \times F \times 1$ dimensional weight matrix. Even though GCN incorporates first order neighborhood, it also enforces parameter sharing between first and zeroth order neighborhood weights, thus fixing its weight matrix size to $C \times F$.

ChebNet: For a K-th order neighborhood model, $W^t$ should have $C
\times F \times (K+1)$ parameters. This is because the model learns a
separate set of $C \times F$ weights for each order degree ranging from $k=0$ to $K$. For example, for a 2nd order model $W^t$ would be a $C \times F \times 3$ dimensional tensor with a forward matrix corresponding to the zeroth, first and second order degrees.


%

\subsubsection{Graph Convolutional Network (GCN)}\
\par This is the first order layer used in the forward models of \cite{kipf-gcn} and \cite{kipf2016variational}. This layer also enforces weight sharing between the first-order and self weights, effectively acting as a regularizer and restricting the model to only one set of parameter weights. Given the adjacency matrix $A$, the filter support is computed in the following steps:

\begin{enumerate}
\item Addition of self-connections: $\tilde{A} = A + I_N$
\item Computation of degree matrix: $\tilde{D} = \sum \tilde{A}$
\item Normalization Step: $ \hat{A} = \tilde{D}^{-\frac{1}{2}}\tilde{A}\tilde{D}^{-\frac{1}{2}} $
\end{enumerate}

\par The normalization step is done to prevent numerical instabilities and is computed as a pre-processing step. Using the normalized adjacency matrix effectively makes averaging the aggregation strategy. The layer-wise model then becomes: $H^{t} = \sigma(\hat{A} H^{t-1} W^{t})$

\par Now considering input $X \in \mathbb{R}^{N \times C}$ with $C$ input channels and convolved output $Z \in \mathbb{R}^{N \times F}$ with $F$ features. The final model with a 2-layer GCN with ReLU non-linearity in hidden layer and softmax in output layer is given as: $Z = f(X, A) = softmax(\hat{A} \ ReLU(\hat{A}XW^{(0)}) \ W^{(1)})$

\par The number of input channels $C$ is equal to the dimensionality of the features of the data (such as the attributes of the node). For example, in case of Cora and Citeseer, it is equal to the size of the bag-of-words feature vector. In case of a featureless approach, $C$ will be equal to the number of nodes in the graph, as each node is then represented with a one-hot representation.

\par Even though the layer-wise model is first-order, stacking multiple layers increases the locality of the filter. For example, a 2-layer GCN would incorporate information in the 2nd order neighborhood of the node for computing the filter. 

\subsubsection{ChebNet}
\par This is the graph convolutional layer originally described in \cite{defferrard2016convolutional}. Unlike the GCN model of Kipf, the Chebnet can incorporate multiple-hop neighborhood information in a single layer. There is also no parameter sharing, as a separate set of parameters are learned for the original node, and each hop neighborhood. For example, a 2nd order Chebyshev layer would have 3 sets of weights and biases in the forward model. The support matrices are computed as Chebyshev Polynomials of the scaled graph Laplacian matrix. The Laplacian matrix is scaled by division with its largest Eigenvalue $\lambda_{max}$ to prevent numerical instabilities when stacking multiple layers. The Chebyshev polynomials themselves are computed recursively as a preprocessing step and the largest Eigenvalue of the laplacian is computed via efficient power iterations.

Given the symmetric normalized graph Laplacian matrix:
\begin{equation}
L = I_n - D^{-\frac{1}{2}}AD^{-\frac{1}{2}},
\end{equation}

\noindent
its rescaled version is given as:
\begin{equation}
\tilde{L} = \frac{2L}{\lambda_{max}} - I_n
\end{equation}
With $T_0 = I$ and $T_1 = \tilde{L}$, the chebyshev polynomials are defined recursively as:
\begin{equation}
T_k(x) = 2xT_{k-1}(x) - T_{k-2}(x).
\end{equation}
Given the rescaled laplacian $\tilde{L}$ and arbitrary model degree K, the filter supports can be pre-computed from $T_0$ to $T_K$ as follows:



\begin{equation}
     \begin{aligned}
& T_0(\tilde{L}) = I \\
& T_1(\tilde{L}) = \tilde{L} \\
&  T_2(\tilde{L}) = 2\tilde{L}^2 - 1 \\
& T_3(\tilde{L}) = 4\tilde{L}^3 - 3\tilde{L} \\
& T_4(\tilde{L}) = 8\tilde{L}^4 - 8\tilde{L}^2 + 1
     \end{aligned}
\end{equation}

The layer-wise model can then be defined as a weighted sum of these components. \\

\begin{equation}
H^{t} = \sigma(\sum_{k=0}^{K}T_k(L)H^{t-1}W_k^t)
\end{equation}

\par Since $\tilde{L}$ is a sparse matrix with $O(|E|)$ elements, the filtering step involves multiplications with $\tilde{L}$ only. The K-th order ChebNet layer will then have K+1 sets of parameter weights and computational complexity $O(|E|)$ where $|E|$ is the number of edges in the graph.

\subsection{Task-Specific Cost Functions}
Here we briefly describe the optimization objective of graph based
learning task used in the experimental evaluation. For the task of
semi-supervised node classification, we use the cross entropy over
all labeled examples.
%
%
The Cross-Entropy loss is used as a supervised training signal for semi-supervised node classification. Given $k$ prediction classes and $Z = GCN(X, A)$ as the matrix of normalized prediction probabilities for each class as computed by the model. The cross entropy loss over all labeled examples $y_L \in Y$ is then given as: $ \mathcal{L} = - \sum_{\ \ell \in y_L} Y_{\ell} \ ln \ Z_{\ell} $

\par For graph convolution models, even though the loss is only computed over labeled examples, the predictions depend on the unlabelled examples as well due to the nature of encoding model. The model is tested via prediction accuracy over a hold-out test set.

%



\subsection{Modularity Optimization Term}
\par The main contribution of this paper is the incorporation of a modularity preserving constraint on the embeddings of the Graph Convolutional Network. This constraint is imposed by the addition of a new loss term into the task-specific cost functions of the previous section.  We can describe the modularity optimization as a secondary objective that is jointly optimized with the primary objectives. As described in the previous section, modularity is a quality function that scores the partition of a given network into $k$ clusters or partitions. Given a cluster assignment matrix $H$ we can simply compute this score as:

\begin{equation}
 Q = tr(\bm{H}^T\bm{B}\bm{H}),
\end{equation}

\noindent
where $ B_{ij} = A_{ij} - \frac{k_i k_j}{2e} $ is the modularity matrix. 

\par Setting the model embeddings as the cluster assignments, we can compute a modularity score for the embeddings using the term given above. Network modularity can then be optimized using gradient descent in any Neural Network pipeline. In this work, we follow the approaches in \cite{weston2012deep} to jointly optimize for both modularity Score and the task-specific in the semi-supervised embedding frame with the graph convolutional model as the deep architecture.

\subsubsection{Regularization Term}
\par We simply subtract the modularity optimization term from the cost function of the output layer of the architecture. The term simply acts as a regularizer that encourages the optimizer to favor weight values that maximize this term. We use a trade-off parameter $\alpha$ to balance between the unsupervised loss and task-specific objectives. Essentially the two objectives share the same layers and model parameters and the model is trained to jointly optimize them. The modified cost function is given as: $ \mathcal{L}_{total} = (1 - \alpha) \mathcal{L}_{supervised}  - \alpha tr(H^T B H ) * (\frac{1}{2e}) $ where $e$ is the number of edges in the graph.

\par The first term of this loss function depends on all labeled examples in the case of node classification. The modularity optimization term however, is computed over all embeddings regardless of the number of training samples. $\alpha$ then becomes a hyper-parameter to be optimized.

\subsubsection{Auxiliary Layer:}
\par An issue with regularizing the output layer of the architecture is that the number of community partitions to be optimized over is fixed to the output layer size. We introduce an auxiliary layer to compute a separate set of representations which dimensionality equal to the number of network partitions to be optimized over.

\par Considering a 2-layer encoder model given by two Graph Convolutional layers denoted by $GCN_1$ and $GCN_2$, the forward model is given as:

\begin{equation}
     \begin{aligned}
& H_{hid} = GCN_1(X,A) \\
& H_{out} = GCN_2(H_{hid}, A)
\end{aligned}
\end{equation}

\par We regularize the intermediate hidden layer representations via an auxiliary feed-forward layer denoted by $MLP_{aux}$, whose input is the hidden layer embeddings and the output is an $n \times k$ cluster assignment matrix $H_{aux}$ where $k$ is the number of partitions.
\begin{equation}
H_{aux} = MLP_{aux}(H_{hid})
\end{equation}

\par The task-specific loss is then evaluated over $H_{out}$ and the modularity optimization term over $H_{aux}$. The total loss is then taken to be a weighted sum of these two.
\begin{equation}
\mathcal{L}_{total} = (1 - \alpha) \mathcal{L}_{supervised}(H_{out}) - \alpha \mathcal{L}_{modularity}(H_{aux})
\end{equation}

\par In this type of architecture, the layers before the branching are shared between the two objectives and receive training signals from both gradients. After branching, the two terms have their own sets of weight parameters to optimize. 

\par At each iteration, the layers unique to the supervised layer (after branching) are only updated via gradients from the supervised loss, and the auxiliary layer only recieves gradient updates from the modularity loss. The layers before branching should recieve a linear combination of the gradients from both output layers (based on $\alpha$).



\section{Experiments}

\subsection{Experimental Setup}

\subsubsection{Evaluation Metrics}
For node classification we use Accuracy score as evaluation metrics. The accuracy score is simply the percentage of correctly classified nodes. 

\subsubsection{Datasets}

\par We also use two bibliographic network datasets, Cora and Citeseer, initially introduced in \cite{sen2008collective}. These possess node features and classification labels for analysis on the semi-supervised classification task. These have been extensively used in the semi-supervised learning literature, including \cite{yang2016revisiting} and \cite{kipf-gcn}. The node features consist of bag-of-word representations of the document text and the edges represent the citation links. The links are assumed to be undirected in this case. 

\subsection{Semi-Supervised Node Classification}

\par Given the network structure, node features and label values of a subset of nodes, the task of semi-supervised node classification is to predict labels for the remaining nodes. The task is different from standard supervised classification because the pair-wise affinity structure of the samples is available as an input. The feature values of both labeled and unlabelled examples can be used for prediction. The basic distinction from pure community detection is that the latter involves inferring topological clusters from structure alone. On the other hand, the labels for node classification have a semantic meaning and might not necessarily correspond to structural communities. However, many information network domains do tend to exhibit correspondence between semantic and topological groups according to the cluster hypothesis, and this can be attributed to the success of many semi-supervised learning methods in recent literature.

\subsubsection{Experiment Design}
\par Since the modularity optimization term provides additional
information about network community structure not available from the
labels themselves, we expect it to be most effective when the number of
labeled examples for training is very small. To verify this, we train
the baseline models and our community-enhanced variants on different
numbers of training examples sampled from the label set. To avoid issues
with label imbalance, we sample an equal number of labels from each
class. We use uniform random sampling in our experiments but other
sampling strategies like PageRank and degree have been explored in
\cite{lin2010semi}. We avoid centrality and degree based sampling
because they tend to return a clustered set of nodes that is not
representative of the full label set. We also sample 1000 addition nodes
as a test set in each run. We do not use a validation set for early
stopping for a fair comparison in the sparse label regime. Instead, we
let each model run for 100 epochs. Since the accuracy is highly
dependent on the initial training sample, we average the results for 20
times with different train-test splits to get a good estimate of the average performance.

\par We test with two Graph Convolutional Encoders, GCN and ChebNet. For
the community-enhancement term, we use two approaches, regularization
via auxiliary layer and direct regularization in the output layer. All
models not counting the auxiliary layers are stacked 2-layer graph
convolutional architectures. This brings us to a total of six different
architecture choices. We test all models for 5, 8, 11, 14, 17 and 20
training labels per class and measure performance for 20 runs. We also
report results for the Iterative Classification Algorithm
\cite{neville2000iterative}, a semi-supervised classification baseline,
averaged over 20 runs. For Cora, the accuracy scores for node
classification for all six neural architectures and the ICA baseline are
detailed in Table \ref{table:cora_acc}. The results for citeseer are omitted because
of space limitations, but they are similar to the results for Cora. We also report the standard error for each instance. Results in the top row of the table represent the baseline methods and the second row shows the proposed architectures in this paper. 


\subsubsection{Results}

\par We note that the community-enhanced ChebNet without the auxiliary layer is the best-performing model on both datasets. ChebNet benefits more from the modularity optimization term because GCN's first-order filter with weight sharing already acts as a regularizer and therefore addition of another unsupervised regularization term has little additional benefit. ChebNet is a more general model with higher number of parameters and can incorporate higher order neighborhood information, allowing it to more easily optimize for secondary tasks. Figures \ref{fig:gcn_cora_acc} to \ref{fig:chebnet_cora_acc} show the relative effect of the community-enhanced loss term on accuracy for both GCN and ChebNet.

\begin{table}\textbf{}
\begin{center}
\caption{Accuracy score and Standard Error on Cora with different number of labeled examples per class, averaged over 20 runs}
\label{table:cora_acc}
\resizebox{\textwidth}{!}{\begin{tabular}{  c c c c c c c  } 
\hline
 Model & 5 & 8 & 11 & 14 & {17} & {20} \\ 
 \hline
ICA & $0.571 \pm 0.013$ & $0.641 \pm 0.011$ & $0.700 \pm 0.010$ & $0.713 \pm 0.007$ & $0.716 \pm 0.007$ & $0.744 \pm 0.007$ \\
GCN & $0.664 \pm 0.012$ & $0.715 \pm 0.010$ & $0.747 \pm 0.007$ & $0.768 \pm 0.005$ & $0.772 \pm 0.006$ & $0.785 \pm 0.004$ \\
ChebNet & $0.612 \pm 0.013$ & $0.685 \pm 0.010$ & $0.732 \pm 0.008$ & $0.735 \pm 0.006$ & $0.762 \pm 0.007$ & $0.783 \pm 0.005$ \\
\hline
GCN-mod & $\bf{0.699 \pm 0.013}$ & $0.738 \pm 0.007$ & $0.755 \pm 0.010$ & $0.775 \pm 0.005$ & $0.791 \pm 0.005$ & $0.786 \pm 0.007$ \\
Chebnet-mod & $0.652 \pm 0.024$ & $\bf{0.745 \pm 0.012}$ & $\bf{0.768 \pm 0.007}$ & $\bf{0.791 \pm 0.004}$ & $\bf{0.813 \pm 0.003}$ & $\bf{0.811 \pm 0.004}$ \\
GCN-aux & $0.670 \pm 0.011$ & $0.703 \pm 0.012$ & $0.748 \pm 0.006$ & $0.754 \pm 0.006$ & $0.780 \pm 0.005$ & $0.775 \pm 0.006$ \\
Chebnet-aux & $0.633 \pm 0.010$ & $0.682 \pm 0.011$ & $0.738 \pm 0.006$ & $0.747 \pm 0.005$ & $0.779 \pm 0.005$ & $0.784 \pm 0.005$ \\
 \hline
\end{tabular}}
\end{center}

\end{table}

\begin{figure}[!htbp]
  \centering
  \begin{minipage}[b]{0.45\textwidth}
  \centering
	\includegraphics[width=\textwidth]{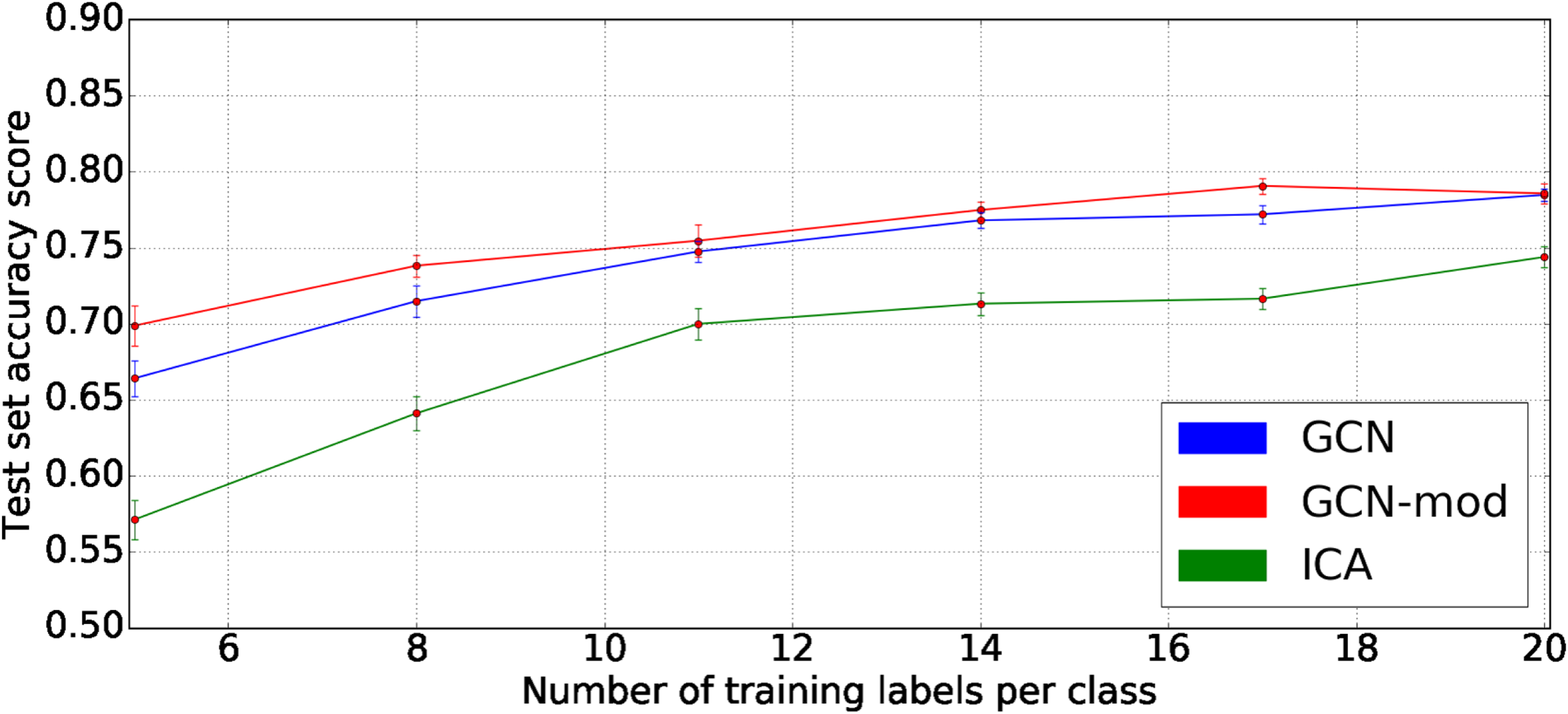}
	\caption{Node Classification Accuracy score for GCN on Cora}
	\label{fig:gcn_cora_acc}
  \end{minipage}
  \hfill
  \begin{minipage}[b]{0.45\textwidth}
  \centering
	\includegraphics[width=\textwidth]{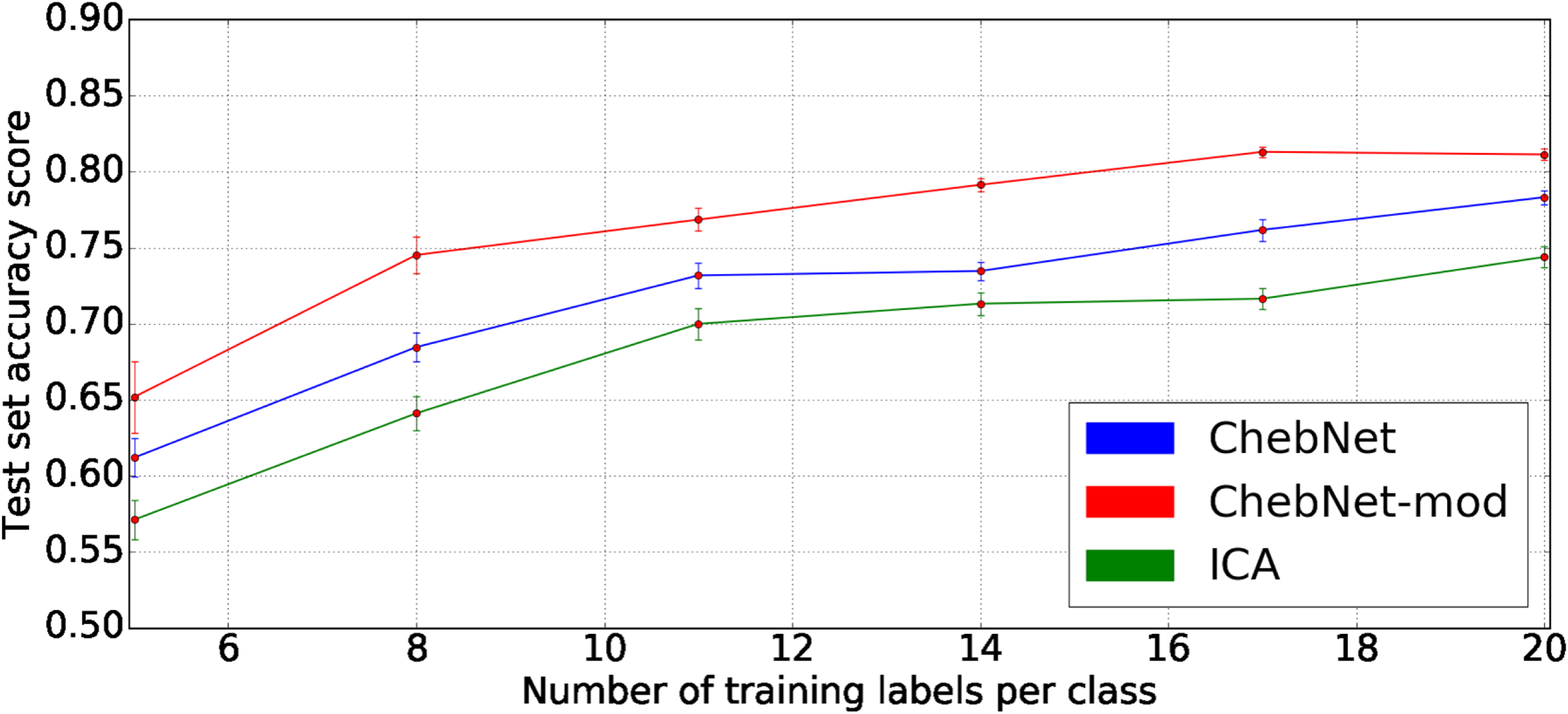}
	\caption{Node Classification Accuracy score for ChebNet on Cora}
	\label{fig:chebnet_cora_acc}
  \end{minipage}
\end{figure}




\subsubsection{Embedding Visualization}
We also visualize the effect of the community-enhancement on the quality of learned model representations in the sparse label regime. We create three instances of a 2-layer ChebNet with values of $\alpha$ set to $0$, $0.5$ and $1.0$ respectively. We train these models on Cora with 5 training labels per class until convergence. We visualize the hidden layer representations of each using T-SNE. These are shown in Figure \ref{fig:mod_emb}. We note that joint optimization with $\alpha = 0.5$ leads to better separation of ground-truth labels as compared to optimizing either term alone. Setting $\alpha = 1.0$ leads to some visible clustering patterns but this does not correspond to a good accuracy score on the given task. We attribute this due to the model converging to local minima of the modularity score loss.

\begin{figure}[!htbp]
  \centering
  \begin{minipage}[s]{0.3\textwidth}
\centering
	\includegraphics[width=\textwidth]{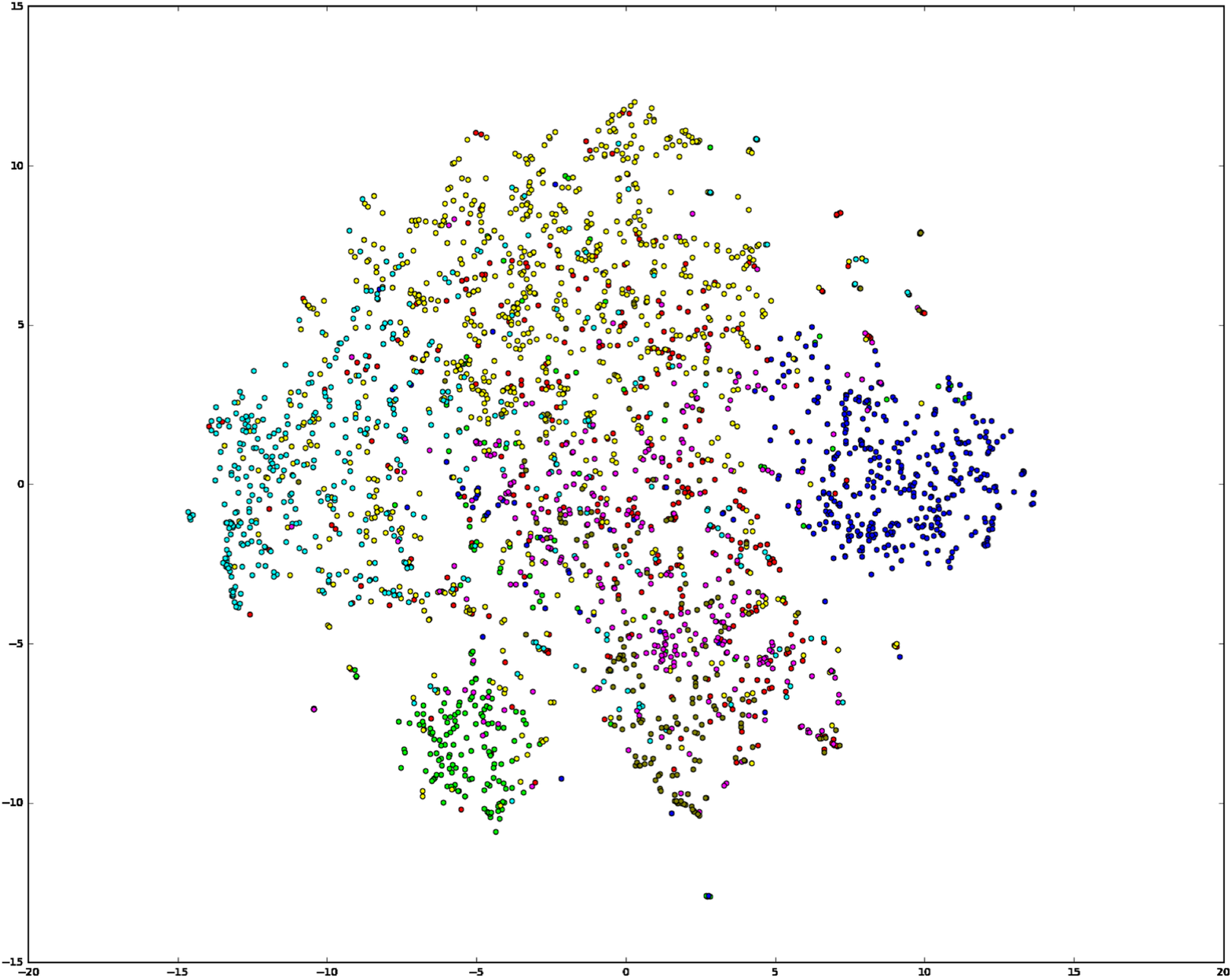}
  \small (a) $\alpha=0.0$

  \end{minipage}
  \begin{minipage}[s]{0.3\textwidth}
\centering
    \includegraphics[width=\textwidth]{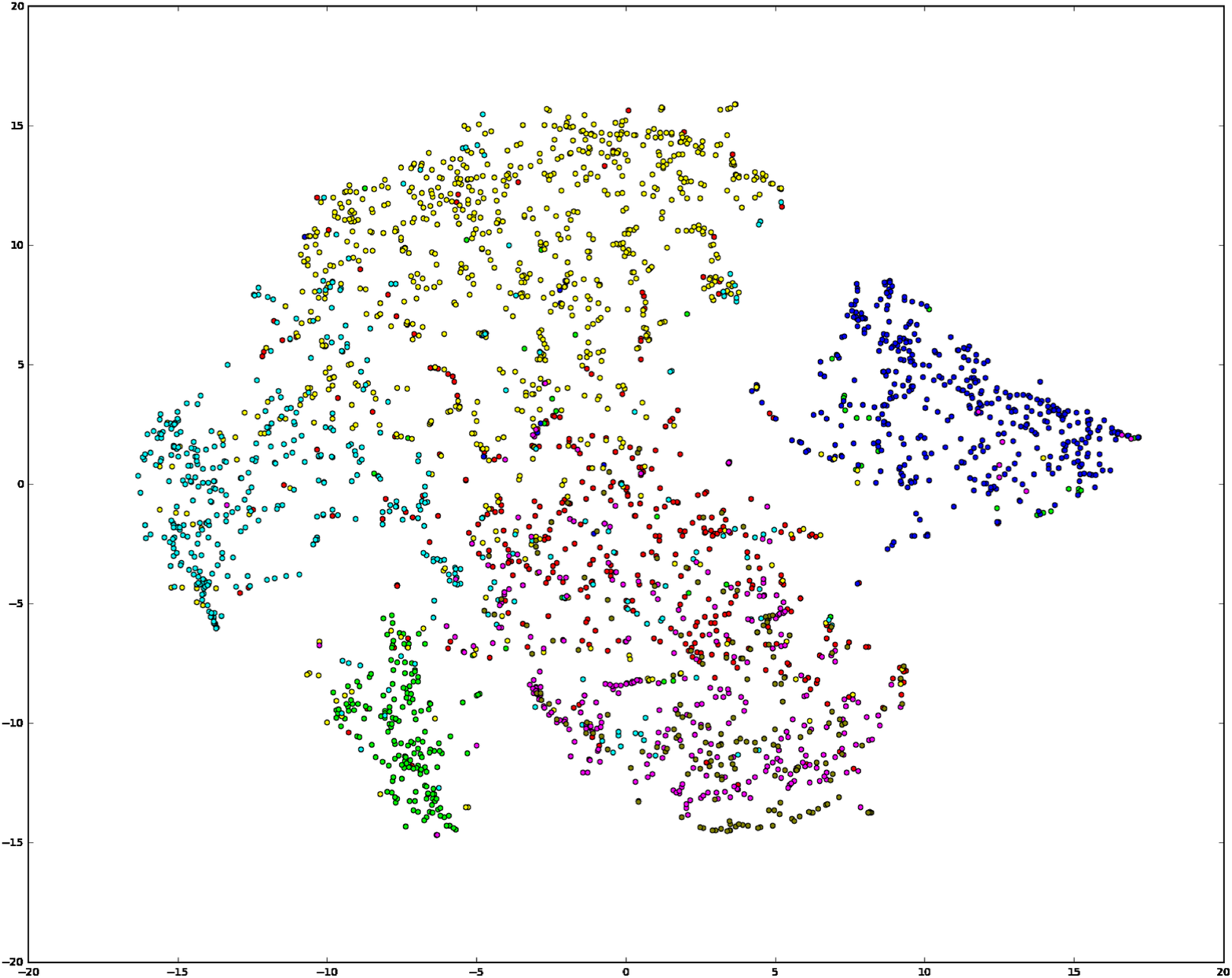}
  \small (b) $\alpha=0.5$

  \end{minipage}
  \begin{minipage}[s]{0.3\textwidth}
\centering
    \includegraphics[width=\textwidth]{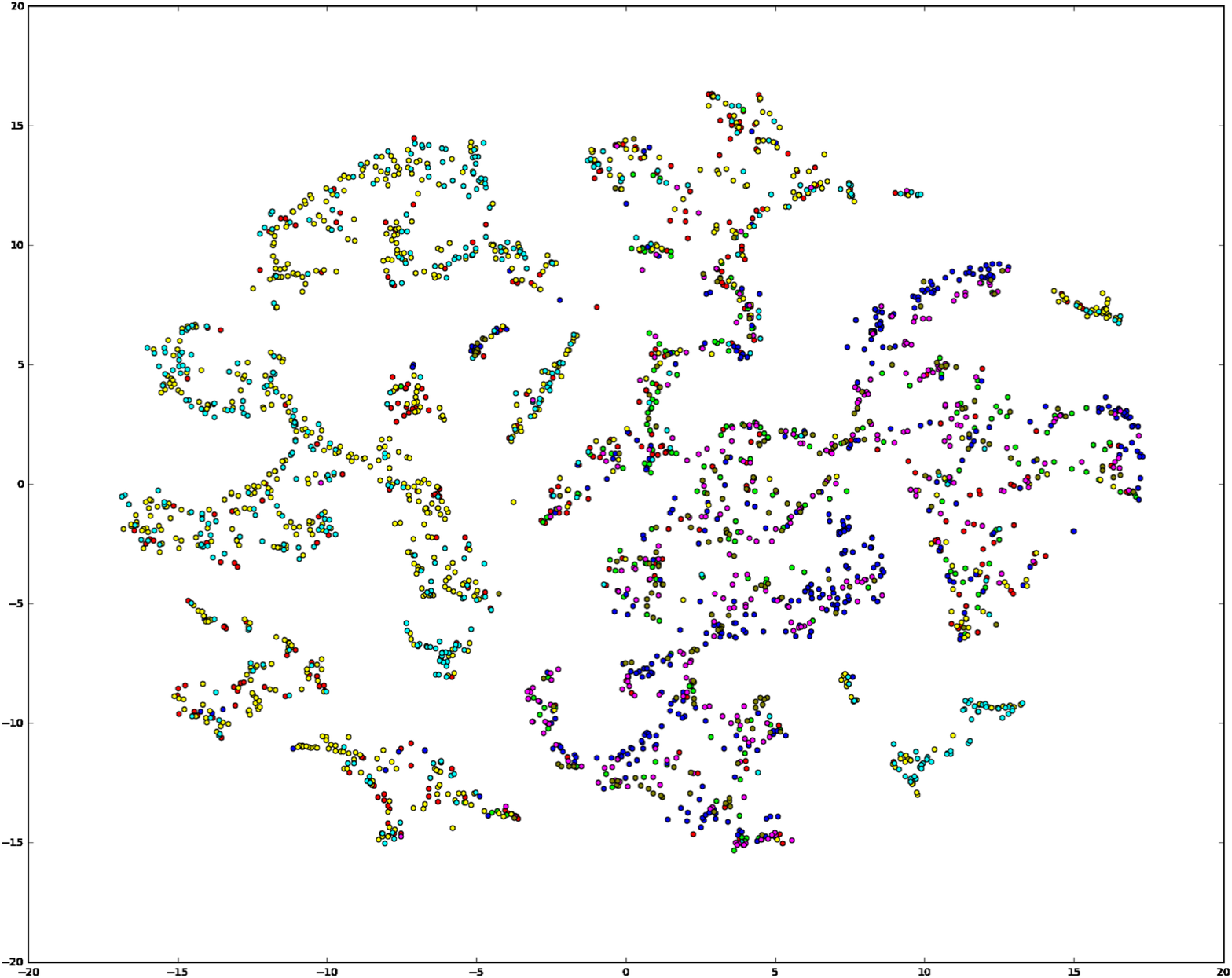}
  \small (c) $\alpha=1.0$

  \end{minipage}
  \caption{T-SNE visualization of hidden layer embeddings of 2nd order 2-layer ChebNet trained on Cora with varying values of $\alpha$}
  \label{fig:mod_emb}
\end{figure}

\section{Conclusion}

\par We successfully incorporate a community structure preserving objective in the graph convolutional semi-supervised learning framework. To the best of our knowledge, this is the first such attempt in this area. We showed that the incorporation of higher level structural information can improve the quality of learned representations for node classification in sparse label regime. We also identified that the specific choice of filter support has a significant impact on the result. Higher-order filters tend to benefit more from the additional network structure preserving terms in the loss function. 


\section*{Acknowledgement}
This work was supported by Tokyo Tech - Fuji Xerox Cooperative Research (Project Code KY260195), JSPS Grant-in-Aid for Scientific Research(B) (Grant Number 17H01785) and JST CREST (Grant Number JPMJCR1687).

\bibliographystyle{plain}
\bibliography{references}
\end{document}